\documentclass{llncs}

\usepackage{graphicx}
\usepackage{amsmath,amssymb} 

\usepackage[nolist]{acronym}
\usepackage{ifthen}
\usepackage[usenames, dvipsnames]{color}

\newboolean{showNotes}
\setboolean{showNotes}{false}

\DeclareRobustCommand{\notesBorrego}[1]{\ifthenelse {\boolean{showNotes}}
 {\textbf{\textcolor{blue}{Jo\~{a}o: #1}}}
 {}
}
\DeclareRobustCommand{\notesRui}[1]{\ifthenelse {\boolean{showNotes}}
 {\textbf{\textcolor{green}{Rui: #1}}}
 {}
}
\DeclareRobustCommand{\notesAtabak}[1]{\ifthenelse {\boolean{showNotes}}
 {\textbf{\textcolor{magenta}{Atabak: #1}}}
 {}
}
\DeclareRobustCommand{\General}[1]{\ifthenelse {\boolean{showNotes}}
 {\textbf{\textcolor{BurntOrange}{General: #1}}}
 {}
}

\newcommand{\customfootnotetext}[2]{{
  \renewcommand{\thefootnote}{#1}
  \footnotetext[0]{#2}}}

\DeclareRobustCommand{\hhref}[1]{\texttt{\url{#1}}}

\begin{acronym}
\acro{COCO}{Common Objects in COntext}
\acro{CoM}{Center of Mass}
\acro{CNN}{Convolutional Neural Network}
\acro{PCA}{Principal Component Analysis}
\acro{SDF}{Simulation Description Files}
\acro{SSD}{Single--Shot Detector}
\acro{YOLO}{You Only Look Once}
\acro{IoU}{Intersection over Union}
\acro{AP}{Average Precision}
\acro{mAP}{Mean Average Precision}
\end{acronym}

\begin{document}
\pagestyle{headings}
\mainmatter

\def\ACCV18SubNumber{***}  


\title{Applying \emph{Domain Randomization} to Synthetic Data for Object Category Detection}

\author{
Jo\~ao Borrego\textsuperscript{$\star$} \and
Atabak Dehban\textsuperscript{$\star$} \and
Rui Figueiredo \and
Plinio Moreno \and
Alexandre Bernardino \and
Jos\'e Santos-Victor
}

\institute{
Instituto Superior T\'ecnico\\
\texttt{\{jborrego,adehban,ruifigueiredo,plinio,alex,jasv\}\\
@isr.tecnico.ulisboa.pt}
}

\maketitle

\customfootnotetext{*}{Authors contributed equally to this manuscript.}


\begin{abstract}
   Recent advances in deep learning--based object detection techniques have revolutionized their applicability in several fields.
   However, since these methods rely on unwieldy and large amounts of data, a common practice is to download models pre-trained on standard datasets and fine-tune them for specific application domains with a small set of domain relevant images.
   In this work, we show that using synthetic datasets that are not necessarily photo-realistic can be a better alternative to simply fine-tune pre-trained networks.
   Specifically, our results show an impressive \textbf{25\% improvement in the \acs{mAP} metric} over a fine-tuning baseline when only about 200 labelled images are available to train.
   Finally, an ablation study of our results is presented to delineate the individual contribution of different components in the randomization pipeline.
\end{abstract}



\section{Introduction}
\label{sec:intro}

With the availability of advanced object detectors~\cite{dai2016r,liu2016ssd,redmon2017yolo9000,ren2015faster}, these systems and their variations have found many applications ranging from face detection~\cite{zhang2017s}, to medical applications~\cite{zhu2018deeplung}, and to robotics~\cite{maiettini2017interactive}.

However, training these systems from scratch is still a challenge as these methods rely on the availability of large, annotated, and high quality datasets.
One common approach to circumvent this issue is to re-use detectors that were pre-trained on large and available datasets such as \ac{COCO}~\cite{lin2014microsoft} and ImageNet~\cite{russakovsky2015imagenet} and later, apply some form of domain adaptation technique \cite{patel2015visual} for the particular task at hand using a smaller, domain specific dataset~\cite{ferguson2017automatic,maeda2018road,van2018inaturalist}.
This approach results in varying degrees of success~(refer to~\cite{zamir2018taskonomy} for a study on how knowledge can be transferred across different tasks).
This line of research has been accelerated, thanks to the availability of high quality open source implementations of \emph{state--of--the--art} object detectors~\cite{Detectron2018,huang2017speed}.

At the heart of these domain adaptation techniques, lies the implicit assumption that there exists some sort of underlying data structure that can be transferred across different domains.
However, this premise does not hold in many applications, specially when the target domain does not significantly overlap with the outdoor images that make up a large portion of both ImageNet and \ac{COCO}.


\begin{figure}[t]
\begin{center}
   \includegraphics[width=0.8\linewidth]{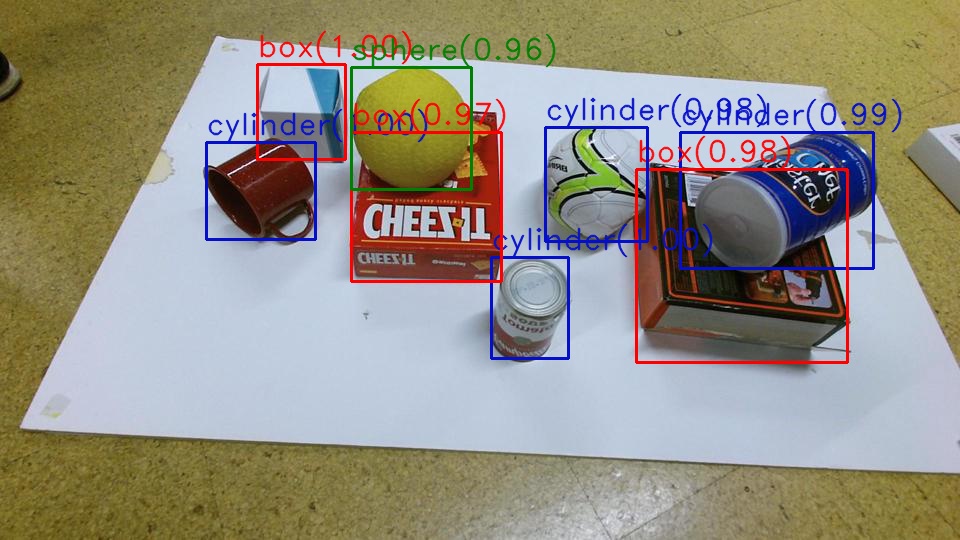}
\end{center}
\caption{
An example image from the test set, annotated by the object detector.
Annotations are red for boxes, blue for cylinders and green for sphere.
}
\label{fig:real_overlay}
\end{figure}


To overcome these challenges, in this work, we are investigating the usage of \emph{domain randomization}~\cite{tobin2017domain} to facilitate the adaptation of an object detector, namely \ac{SSD}~\cite{liu2016ssd}, to detect three classes of objects: cylinders, spheres, and boxes.

This task is accomplished using an open source plugin \cite{borrego2018generic} developed for Gazebo simulator~\cite{koenig2004design}.
This plugin was selected as it streamlines the generation and rendering of different objects as long as their mesh description is available.
In addition, adding parametric classes of objects using this plugin is quite straightforward.
Finally, Gazebo is the current de facto standard for robotics which covers several physics engines, families of robots, different type of actuators\footnote{\hhref{https://en.wikipedia.org/wiki/Robotics\_simulator} as of \today}. We believe that roboticists will build upon these features and implement \emph{domain randomization} experiments where multiple robots may interact with several objects while learning new skills.

According to our experiments, domain randomization can substantially increase the accuracy of object detectors at least in situations where only a relatively small domain--specific dataset of annotated images is available.
Even though not completely generalizable, the fact that the synthetic dataset does not necessarily need to be photo-realistic helps to significantly lower the barrier in applying this technique in different applications.

\

The main contributions of this paper can be summarized as following:
\begin{itemize}
    \item We have shown substantial improvements in the accuracy of \ac{SSD} compared to the case where it was simply fine-tuned on a small, domain--specific dataset;
    \item We conducted a comprehensive study in order to determine the contribution of individual components of the pipeline and discuss the importance of viewpoint variations, different types of textures and number of available images for training;
    \item We have made significant contributions to an open source Gazebo plugin, which has resulted in \emph{doubling} the speed of scene generation pipeline, by effectively removing redundant object load times. These modifications have greatly facilitated the study of domain randomization in object category detection.
\end{itemize}

The rest of this paper is organized as follows: in section~\ref{sec:related-work} we examine the related work on different domain adaptation techniques relevant to object detection, that have been studied in the literature.
In section~\ref{sec:methods} we explain the setup of the experiments, as well as our contributions to the Gazebo plugin which has made this work possible.
For the sake of completeness, a brief overview of \ac{SSD} is also provided.
Section~\ref{sec:results} discusses the results of using domain randomization on object detection and the significance of different components in the domain randomization pipeline.
It also benchmarks the importance of our contributions to the Gazebo plugin for scene generation.
Finally, we draw our conclusions and discuss promising future research directions in section~\ref{sec:conclusions}.


\section{Related Work}
\label{sec:related-work}

Recent advances on deep learning and parallel computing have boosted research and many breakthroughs in machine learning and computer vision.
Being capable of learning the underlying highly nonlinear structure of high dimensional data, they have achieved \textit{state--of--the--art} performance in image classification~\cite{rawat2017deep}, detection~\cite{ren2015faster} and segmentation tasks~\cite{he2017mask}.
However, supervised training of deep neural networks relies on the availability of large datasets, hand-labeled in a laborious and time consuming manner.

In this section we overview the main concepts and related work on automated, computer driven data augmentation techniques for computer vision applications.

\subsection{Reality Gap}

The discrepancy between the real world and simulated, computer generated environments is often referred to as the reality gap.
There are two common approaches to bridge this disparity: either reducing the gap by attempting to increase the resemblance between the two domains or explore methods that are trained in a more generic domain, representative of both simulation and reality domains simultaneously.
To achieve the former, one may increase the accuracy of the simulators in an attempt to obtain high-fidelity results~\cite{johnson2017driving,zhu2017target}; or use Generative Adversarial Networks (GANs) to turn simulated images more photo-realistic~\cite{shrivastava2017learning}.
Both methods require great effort in the creation of systems which model complex physical phenomena to attain realistic simulation.
Our work focuses mainly on the second approach.
Instead of diminishing the reality gap in order to use traditional machine-learning methods, we analyze methods that are aware of this disparity.

\subsection{Data Augmentation}

 An alternative approach to obtain large amounts of annotated training data is to enrich a small dataset with new labelled elements.
 In ~\cite{Georgakis2016}, the authors generate synthetic composite images for training neural networks for object detection.
They propose methods in which 2D cropped object images are superimposed into a real-world RGB-D scene.
Moreover, their proposal integrates scene contextual information in the data generation process.
 
This work demonstrates that the performance of state-of-the-art object detectors performed better when trained with both synthetic and real data than with real data alone.
The data generation method is tested with two publicly available datasets, GMU-Kitchens~\cite{Datasets:GMU_Kitchens} and Washington RGB-D Scenes V2~\cite{Datasets:Washington}.

\subsection{Domain Randomization}

Rather than attempting to perfectly emulate reality, we may create models that strive to achieve robustness to high variability in the environment.
Domain randomization is a simple yet powerful technique for generating training data for machine-learning algorithms.
The goal is to synthetically generate or enhance the data, in order to introduce random variances in the environment properties \emph{that are not essential to the learning task}.
 This idea dates back to at least 1997~\cite{Jakobi97}, with Jakobi's observation that evolved controllers exploit the unrealistic details of flawed simulators.
 His work on evolutionary robotics studies the hypothesis that controllers can evolve to become more robust by introducing random noise in all the aspects of simulation which do not have a basis in reality, and only slightly randomizing the remaining which do.

It is expected that given enough variability in the simulation, the transition between simulated and real domains is perceived by the model as a mere disturbance, to which it has became robust.

Concurrent to our work,~\cite{tremblay2018training} reports the effect of overlaying real textures on the accuracy of state-of-the-art object detectors in a single-class outdoor car detection scenario. In contrast, we report the impact of overlaying synthetically generated patterns with different characteristics and increasing complexity on the accuracy metrics in a multiple-class indoor detection of parametric shape primitives scenario.


\section{Methods}
\label{sec:methods}

In order to apply an object detector in a new domain, it is necessary to collect some training samples from the domain at hand.
Labelling data for object detection is harder than labelling it for object classification, as bounding box coordinates are needed in addition to target object's identity, which adds to the importance of optimally benefiting from the available data. 

After data collection, a detector is selected, commonly based on a trade-off between speed and accuracy, and is fine-tuned using the available ``target domain'' data.
Our proposal is to use a synthetic dataset, with algorithmic variations in irrelevant aspects of objects of interest, instead of relying on pre-trained networks on datasets which share little resemblance to the task at hand.
This approach is further detailed in this section.

\subsection{\acl{SSD}}

In all of our experiments, \ac{SSD} was used as the base detector as it is one of the few detectors that can be applied in real-time while showing a decent accuracy. However, we expect our results to directly generalize to other deep learning based detectors.

The inner workings of \ac{SSD} is briefly described here, however, readers should refer to the original publication~\cite{liu2016ssd} for a comprehensive study of the detector.

At the root of all deep learning based object detectors, there exists a base \ac{CNN} which is used as feature extractors for further down-stream tasks, \emph{i.e.}~bounding box generation and foreground/background recognition.
Similar to \acs{YOLO} architecture~\cite{redmon2017yolo9000}, \ac{SSD} takes advantage from the concept of \emph{priors} or \emph{default boxes}\footnote{Called \emph{anchor box} in \acs{YOLO}.} where each cell identifies itself as including an object or not, and where this object exists, relative to a default location.
However, unlike \acs{YOLO}, \ac{SSD} does this at different layers of the base \ac{CNN}.
Since neurons in different layers of \acp{CNN} have different receptive fields in terms of size and aspect ratios, effectively, objects of various shapes can be detected.

During training, if a ground truth bounding box matches a default box, \emph{i.e.}~they have an \ac{IoU} of more than $0.5$, the parameters of how to move this box to perfectly match the ground truth are learned by minimizing an smooth L1 metric.
Hard negative mining is used to create a more balanced dataset between foreground and background boxes.
Finally, Non-Maximum Suppression (NMS) is used to determine the final location of the objects.

Unlike the original \ac{SSD} architecture, we used MobileNet~\cite{howard2017mobilenets} as the base \ac{CNN} for feature extraction in all experiments.
MobileNet changes the connections in a conventional \ac{CNN} to drastically reduce its number of parameters, without having a significant toll on performance, relative to a comparable architecture.

\subsection{Contributions to Gazebo plugin}

Our contribution to the open-source Gazebo plugin for domain randomization~\cite{borrego2018generic} consists of an optimization in the scene composition, which almost doubled performance.
Originally, each scene required parametric objects to be generated from a \ac{SDF}\footnote{\hhref{http://sdformat.org/} as of \today} formatted string, which was altered during run-time in order to allow for different object dimensions and visuals.
Furthermore, objects were created and destroyed in between scenes.

Instead, we first spawn the maximum number of objects on scene of each type.
Then, in each scene we alter their visual properties from within the Gazebo engine, by for instance changing their scale and pose,
which results in a substantial performance boost.


In addition, we improved the existing auxiliary texture generation module in order to exploit parallelism in Perlin noise generation, using OpenMP\footnote{\hhref{https://www.openmp.org/} as of \today} framework.

\subsection{Experiment design and setup}
\label{subsec:design-and-setup}

We have conducted various experiments and tests to quantify the results of different scenarios.
Initially, two sets of 30$k$ synthetic images are generated.
The modifications we mentioned in the previous subsection have greatly facilitated this process.
These two sets differed from one another by the degree in which the virtual camera in the scene has changed its location.
In the first set, the viewpoint was fixed, whereas in the other set, its location varied largely across the scene.
More details will be provided in section~\ref{sec:results}.

\begin{figure}[!ht]
\centering
   \includegraphics[width=0.55\linewidth]{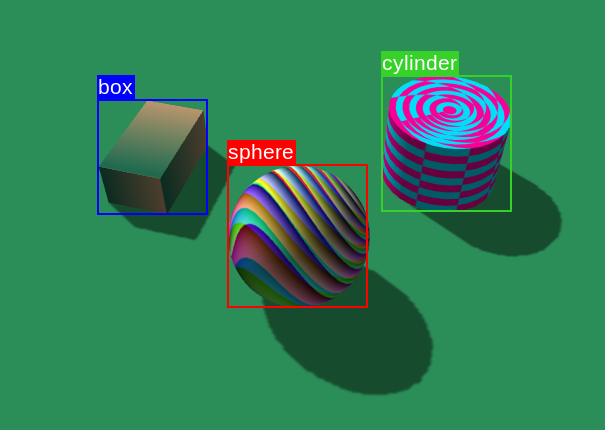}
\caption{
Example synthetic scene employing all 4 texture patterns.
Labelled by the plug-in.
The ground has a flat color, box has gradient, cylinder has chess and sphere has Perlin noise.
}
\label{fig:textures}
\end{figure}

Four types of textures were used in the generation of synthetic images, which have been employed in recent research applying domain randomization~\cite{tobin2017domain,james2017transferring}.
Specifically, these include flat colors, gradients of colors, chess patterns, and Perlin noise~\cite{Perlin02}, which can be seen in Figure~\ref{fig:textures}.

In addition, we have collected 250 real images in the lab, out of which 49 contain objects unseen in training, for the sole purpose of reporting final performance~(Fig.~\ref{fig:real_overlay}). 
The train, validation and test partitions of our real image dataset is specified in Table~\ref{tab:real_dataset}.

\begin{table}[!ht]
\centering
\caption{Number of real images in train, validation and test partitions.}
\label{tab:real_dataset}
\begin{tabular}{llll}
    \hline\noalign{\smallskip}
    Training  $\qquad\qquad$ & Validation  $\qquad\qquad$ & Test  $\qquad\qquad$ & Total $\qquad\qquad$ \\
    \hline\noalign{\smallskip}
    175      & 26         & 49   & 250 \\
    \hline
\end{tabular}
\end{table}

In this dataset, there was no consideration to explicitly keep the percentage of different classes balanced~(Table~\ref{tab:imbalanced}), as such, we have also reported precision-recall curves for each class.
Finally, all our reported metrics are calculated with an \ac{IoU} of $0.5$.

\begin{table}[!ht]
\centering
\caption{Percentage of different classes in the real dataset.}
\label{tab:imbalanced}
\begin{tabular}{lllll}
\hline\noalign{\smallskip}
Partition $\qquad$ &
\# Box $\qquad$ &
\# Cylinder $\qquad$ &
\# Sphere $\qquad$ &
Total $\qquad$ \\
\hline\noalign{\smallskip}
Train set & 502 (63\%) & 209 (26\%)  & 86 (11\%) & 797 \\
Test  set & 106 (40\%) & 104 (40\%)  & 53 (20\%) & 263 \\
\hline
\end{tabular}
\end{table}

For baseline calculations, we have used \ac{SSD}, trained on \ac{COCO}, and fine-tuned it on the train set until the performance by validation set failed to improve.

In other experiments, we have used MobileNet which was trained on ImageNet as the \ac{CNN} classifier of \ac{SSD} and first fine-tuned it on synthetic datasets with bigger learning rates and later, in some experiments, fine-tuned again with smaller learning rates on the real dataset.

Finally, smaller synthetic datasets of $6k$ images were generated, each with a type of texture missing, and an additional baseline for comparison, which includes every pattern type.
These datasets allowed us to study the contribution of each individual texture in the final performance, as well as performance comparison of the smaller synthetic datasets.

\section{Experiments and results}
\label{sec:results}

\begin{figure}[!ht]
\centering
    \begin{minipage}[b]{0.48\textwidth}
        \centering
        \includegraphics[width=\textwidth]{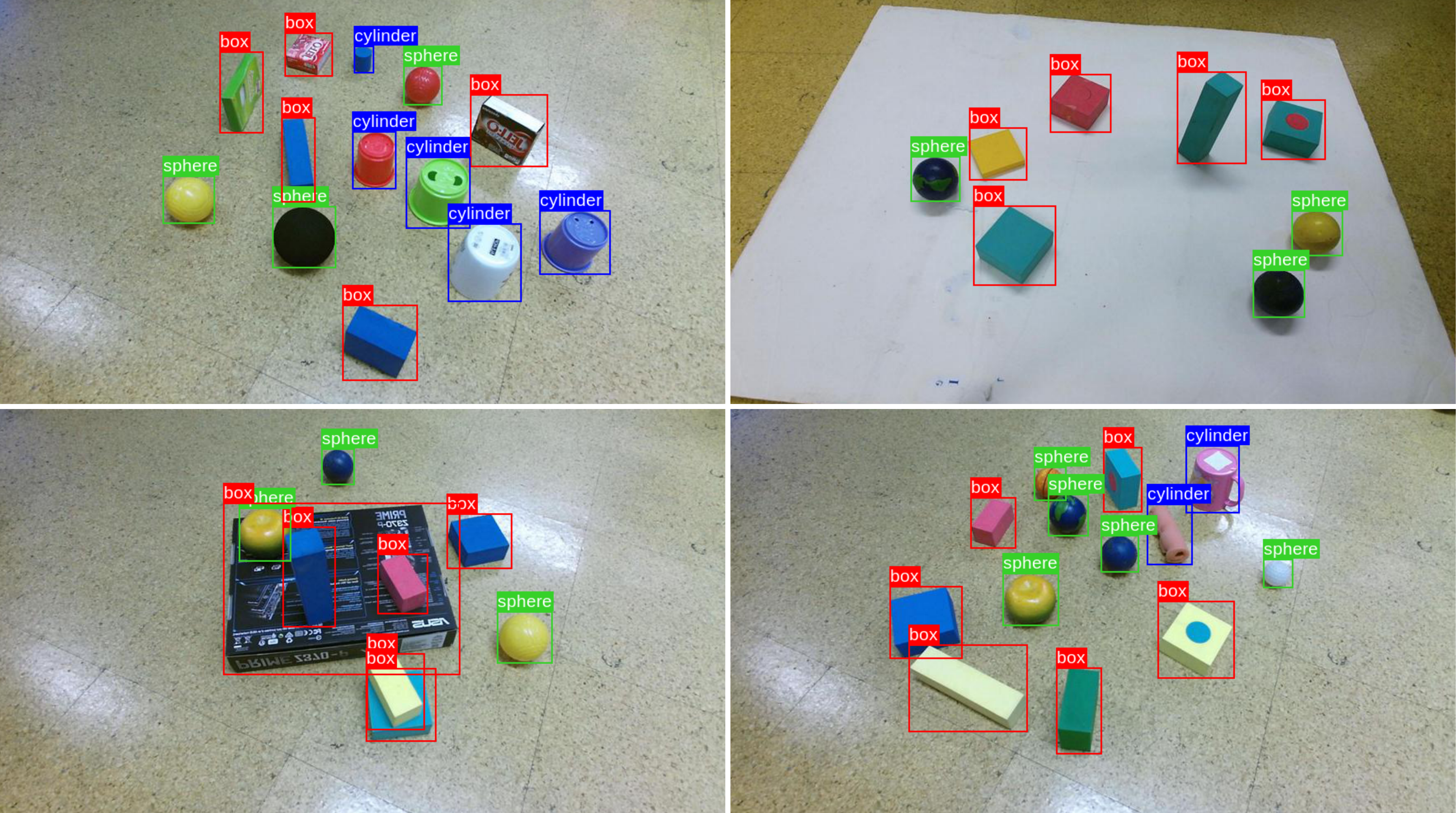}
        {\textbf{(a)} Training Set Examples;}
        \label{fig:train_img}
    \end{minipage}
    ~
    \begin{minipage}[b]{0.48\textwidth}
        \centering
    \includegraphics[width=\textwidth]{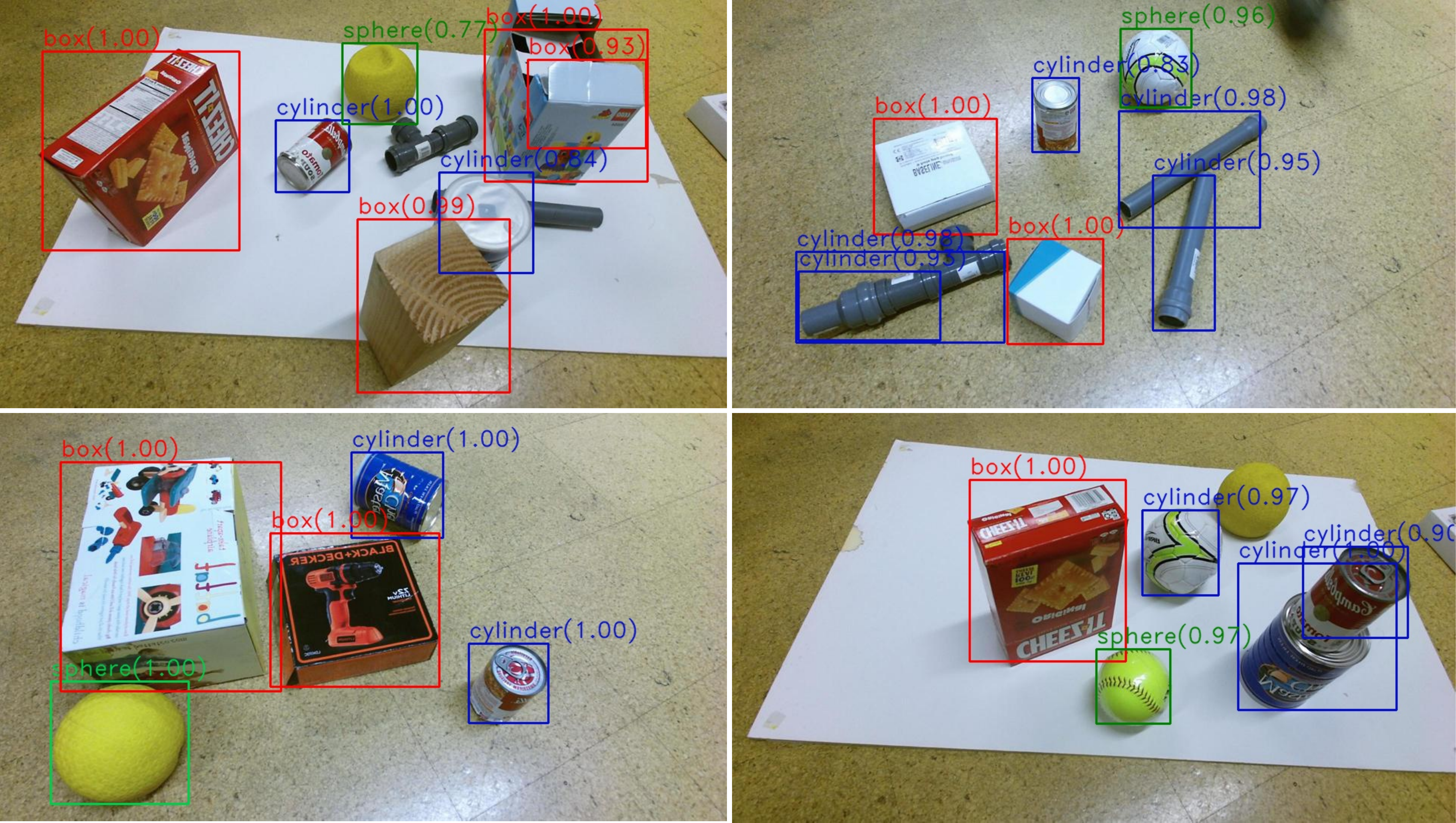}
        {\textbf{(b)} Test Set Examples;}
        \label{fig:test_img}
    \end{minipage}
    \caption{
    Example images from real~\textbf{(a)}~training and~\textbf{(b)}~test sets, annotated with ground truth and detector outputs, respectively.
    }
    \label{fig:train_test_img}
\end{figure}

All synthetic images have Full-HD (1920 $\times$ 1080) resolution and are encoded in \texttt{JPEG} lossy format, to match the training images taken by Kinect v2.0 that were used in our experiments.
For training and testing, images are down-scaled to half these dimensions (960 $\times$ 540) which is the resolution employed for all test scenarios in our pipeline.
Examples of the real datasets can be seen in Fig.~\ref{fig:train_test_img}.

Networks were trained with mini-batches of size $8$, on a machine with two Nvidia Titan Xp GPUs, for a duration depending on the performance in a real image validation set.
We have only used horizontal flips and random crops, with parameters reported in original \ac{SSD} paper, as the pre-processing step, since we are interested in studying the effects of synthetic data and not different pre-processings.
Finally, in compliance with the findings in~\cite{tremblay2018training}, all the weights of the network are being updated in our experiments.

Our code and dataset are currently hosted on GitHub\footnote{\hhref{https://github.com/jsbruglie/tf-shape-detection}, as of \today} and our Laboratory's webpage\footnote{\hhref{http://vislab.isr.ist.utl.pt/datasets/\#shapes2018}, as of \today}.

\subsection{Benchmarking contributions to Gazebo plugin}

In \cite{borrego2018generic}, the authors state that a dataset of 9.000 Full-HD (1920 $\times$ 1080) images took roughly 3 hours to generate.
In a similar computer, we tested the plugin with our modifications and obtained almost double of the speed performance, generating 9.000 synthetic images in little over 1h30min, albeit resorting to a larger set of available random textures (a total of 60.000 textures, compared to the reported 20.000), which expectedly should have increased run-time.

Our novel approach allows us to alter the properties of the objects directly through the rendering engine API, which is much more efficient than spawning and removing objects with different features.
Specifically, objects are spawned below the ground plane and moved to desired location in the new scene.
By changing its scale vector we can effectively morph the object shape.
Finally, we load the random textures as Gazebo resources on launch, and can apply them directly, although they are only loaded into memory once they are required by the rendering engine.

\subsection{Effects of domain randomization on object detection}

In this subsection, we wish to quantify how much an object detector performance would improve due to the usage of synthetic data.
To this purpose, initially, we fine-tuned a \ac{SSD}, pre-trained on \ac{COCO} dataset with our real image dataset for 16.000 epochs, which we determined to be sufficient by evaluating the performance on our validation set.
We used a decaying learning rate $\alpha_0=0.004$, with a decay factor $k=0.95$ every $t=100k$ steps. 
In the subsequent sections we refer to this network as baseline. 

Afterwards, we trained a \ac{SSD} with only its classifier pre-trained on ImageNet, using our two synthetic datasets of $30k$ images each, as described in section~\ref{subsec:design-and-setup}.

Both of these datasets contain simulated tabletop scenarios with a random number of objects $N \in [2,7]$, each in one of three classes: box, cylinder or sphere.
These objects are placed randomly on the ground plane in a $3\times3$ grid, to avoid overlap.

\begin{figure}[!ht]
\centering
    \begin{minipage}[b]{0.4\textwidth}
        \centering
        \includegraphics[width=\textwidth]{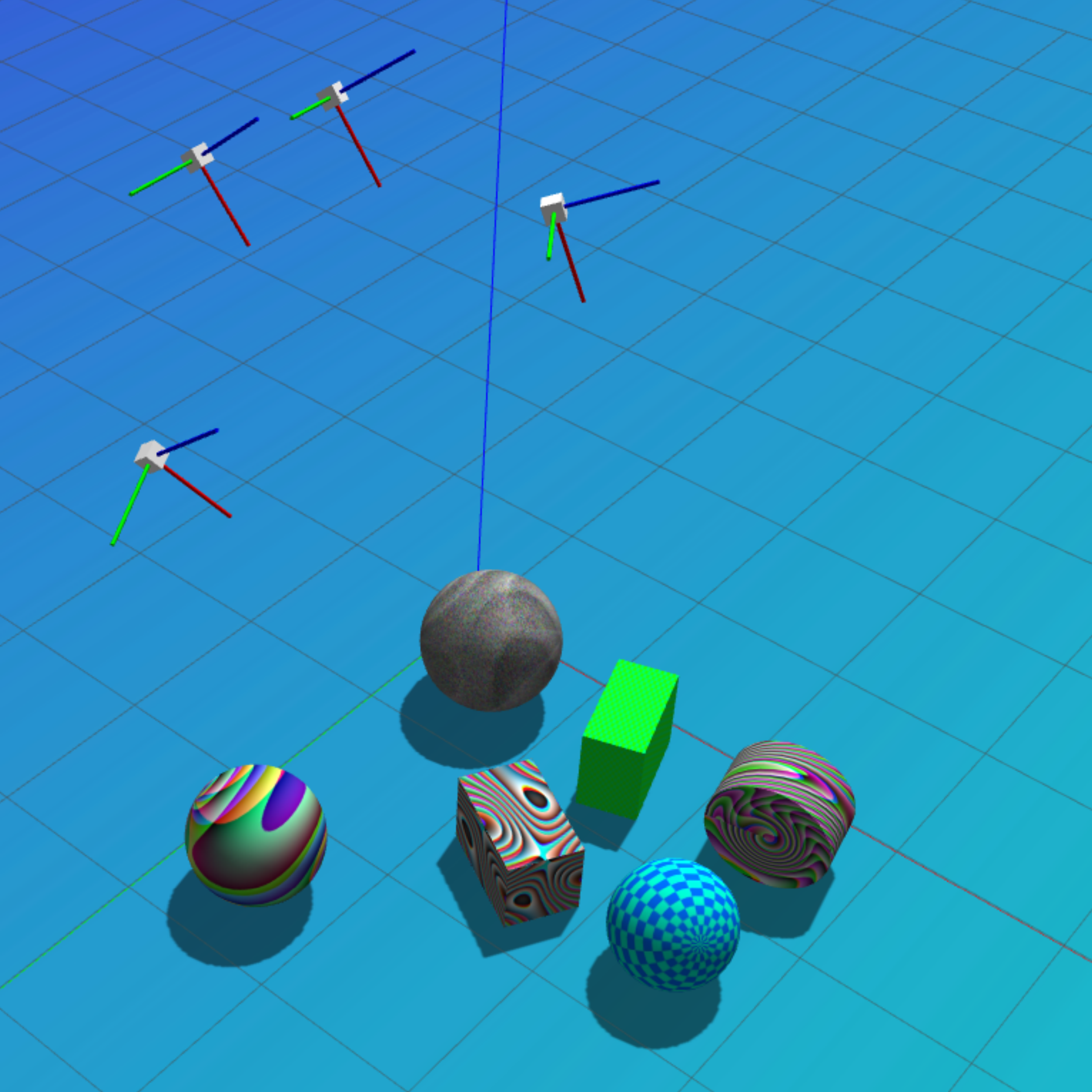}
        {\textbf{(a)} Moving Viewpoint;}
        \label{fig:mv}
    \end{minipage}
    ~
    \begin{minipage}[b]{0.4\textwidth}
        \centering
        \includegraphics[width=\textwidth]{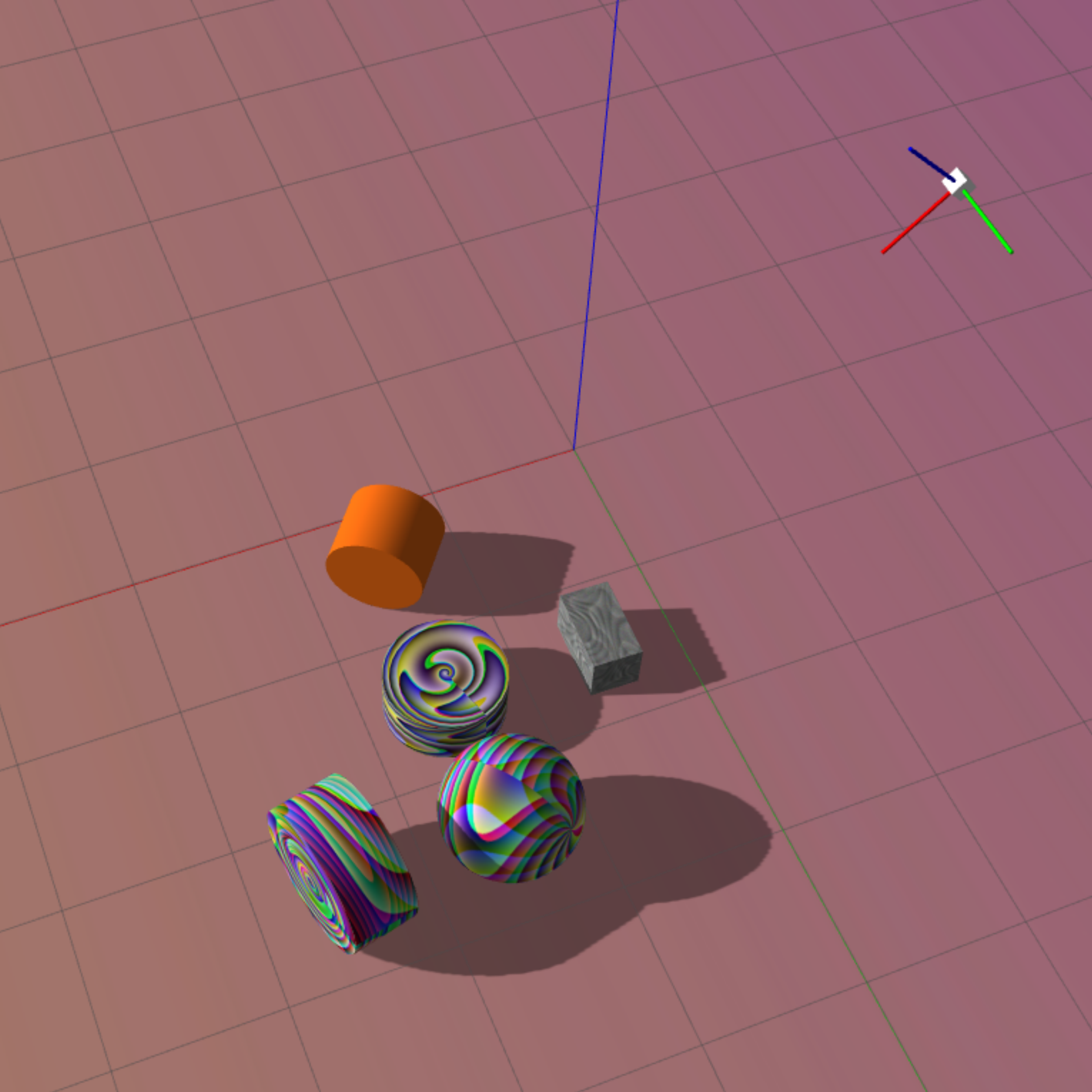}
        {\textbf{(b)} Fixed Viewpoint;}
        \label{fig:fv}
    \end{minipage}
    \caption{
    Viewpoint candidates in synthetic scene generation.
    Left: Viewpoint changes both position and rotation in between scenes.
    Subfigure represents four possible camera poses.
    Right: Viewpoint is static.
    }
    \label{fig:viewpoints}
\end{figure}

In the first dataset, the camera pose is randomly generated for each scene, such that it points to the center of the object grid.
This generally results in high variability in the output, which may improve generalization capabilities of the network at the expense of added difficulty to the learning task, as, for instance, it exhibits higher levels of occlusion.
In the second dataset, the camera is fixed overlooking the scene at a downward angle, which is closer to the scenario we considered in the real dataset.
Example scenes with viewpoint candidates for each dataset are shown in Figure~\ref{fig:viewpoints}.

In addition to the camera, the scene light source is always allowed to move in a manner akin to the camera, in the first dataset. 

Similar to the baseline, the networks were trained on these datasets for over 90 epochs, based on their performance on the validation set employing an exponentially decaying learning rate, starting at $\alpha_0=8\times10^{-3}$, and a decay of $k=0.95$ every $t=50k$ steps. 
These networks were then directly applied to the test set~(which has real images) without any fine-tuning on our dataset of real object data, in order to quantify how much knowledge can be directly transferred from synthetic to real domain.

\begin{figure}[!ht]
\centering
   \includegraphics[width=0.7\linewidth]{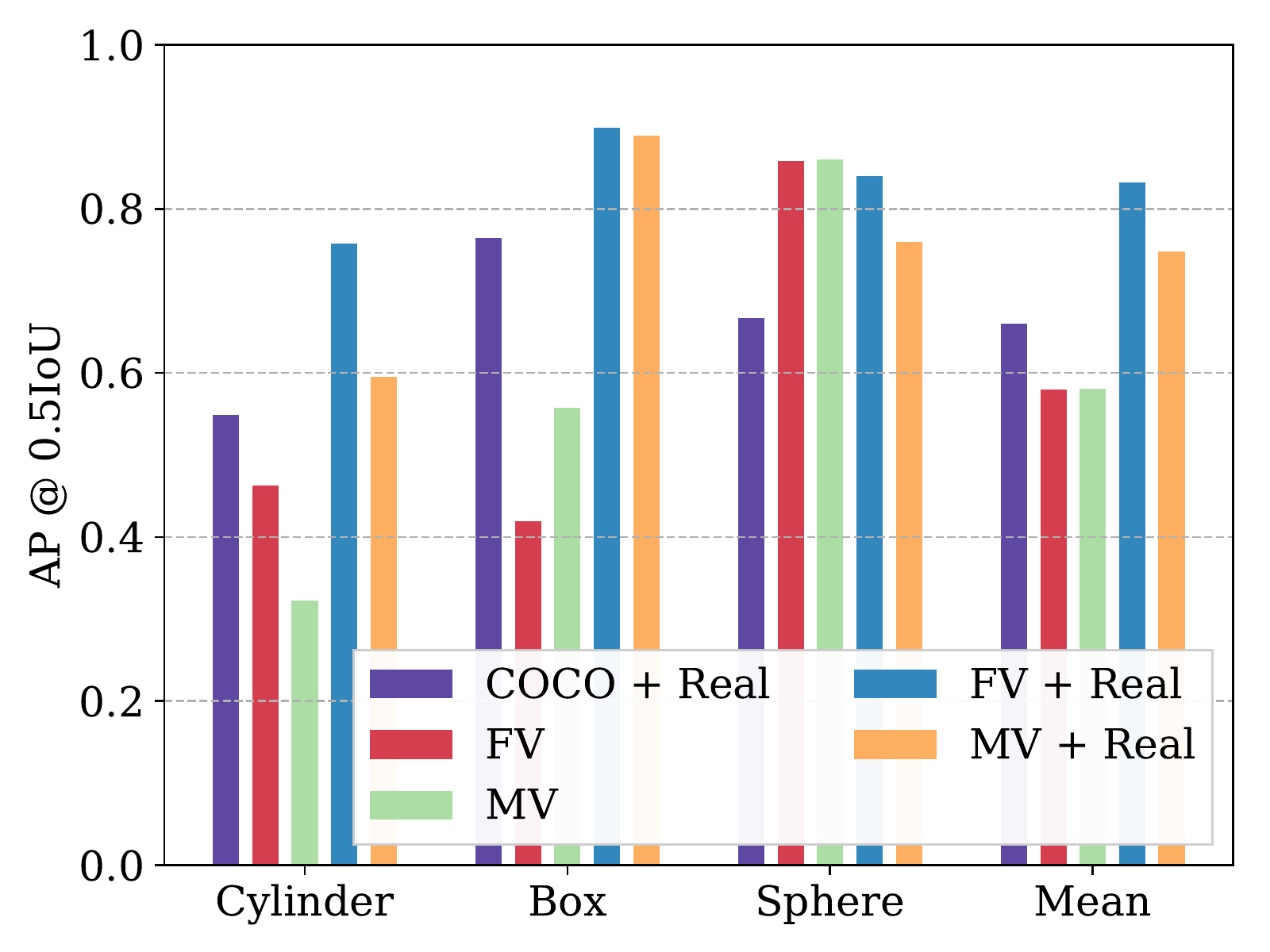}
\caption{
Per class \acs{AP} and \ac{mAP} of different detectors.
MV:~Moving Viewpoint;
FV:~Fixed Viewpoint;
Real:~fine-tuned on the real dataset
}
\label{fig:ap_30k}
\end{figure}

Finally, these two detectors were fine-tuned on the real dataset for over 2200 epochs and with a fixed learning rate of $\alpha=10^{-3}$.
The result of this analysis is depicted in Fig.~\ref{fig:ap_30k} and summarized in Table~\ref{tab:ap_30k}.

\begin{table}[!ht]

\centering

\caption{
\ac{SSD} performance on test set.
For abbreviations refer to Fig.~\ref{fig:ap_30k}.
}
\label{tab:ap_30k}
\begin{tabular}{lllll}
\hline\noalign{\smallskip}
Run $\qquad$ &
mAP $\qquad$ &
AP Box $\qquad$ &
AP Cylinder $\qquad$ &
AP Sphere $\qquad$ \\
\hline\noalign{\smallskip}
COCO + Real $\qquad$ & 0.6598 & 0.7640 & 0.5491 & 0.6664 \\
FV          & 0.5801 & 0.4190 & 0.4632 & 0.8581 \\
MV          & 0.5804 & 0.5578 & 0.3230 & \textbf{0.8603} \\
FV + Real   & \textbf{0.8319} & \textbf{0.8988} & \textbf{0.7573} & 0.8395 \\
MV + Real   & 0.7480 & 0.8896 & 0.5954 & 0.7591 \\
\hline
\end{tabular}

\end{table}

The network trained on the dataset with no camera pose variation and fine-tuning on real data exhibits the best performance at $0.83$ \ac{mAP}, which corresponds to an improvement of $\mathbf{26\%}$ over baseline.

Furthermore, we can observe that although the networks trained only on each of the synthetic datasets obtain similar \ac{mAP} values in the test set (roughly $0.58$), their results differ greatly after fine-tuning on real images.
This observation suggests that the changes in camera pose, seen in the synthetic dataset have indeed hurt the performance, as our test set does not exhibit these variations.
However, it is expected that the network trained on the corresponding dataset is more robust and would perform better if it was tested against a dataset with varying camera/light positions.

\begin{figure}[!ht]
\centering
    \begin{minipage}[b]{0.4\textwidth}
        \centering
        \includegraphics[width=\textwidth]{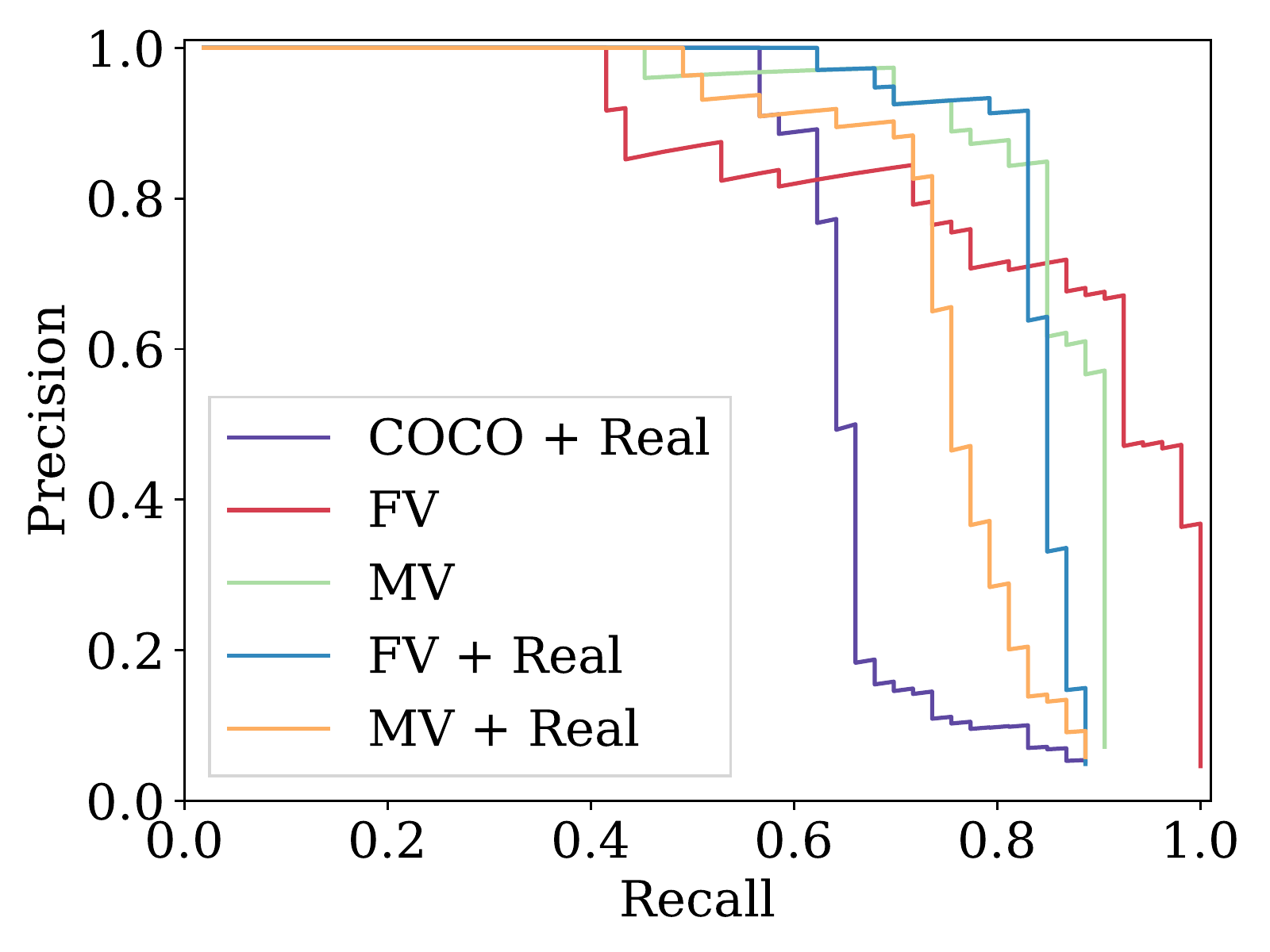}
        {\textbf{(a)} Sphere;}
        \label{fig:pr_30k_sphere}
    \end{minipage}
    ~
    \begin{minipage}[b]{0.4\textwidth}
        \centering
        \includegraphics[width=\textwidth]{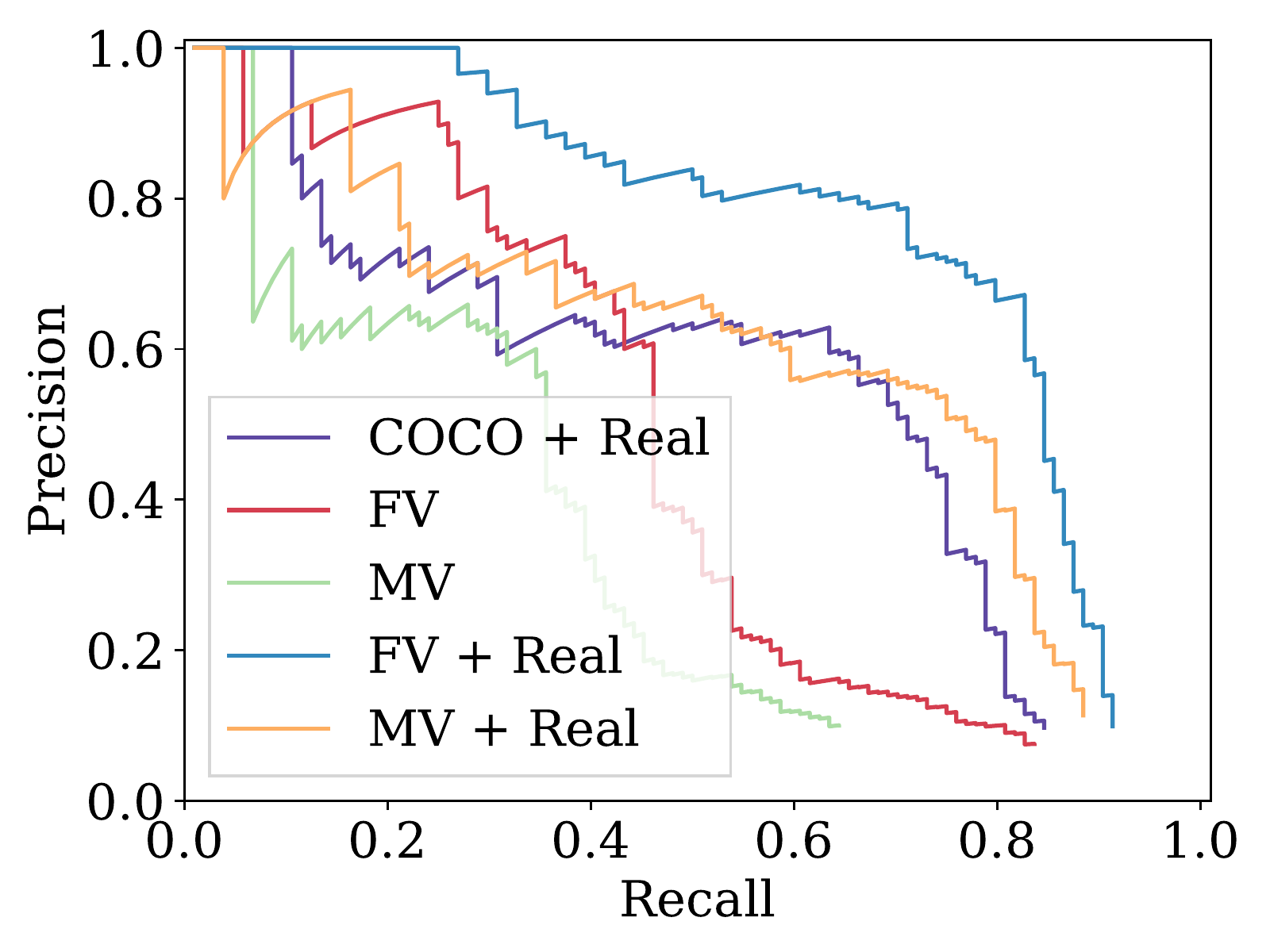}
        {\textbf{(b)} Cylinder;}
        \label{fig:pr_30k_cylinder}
    \end{minipage}
    ~
    \begin{minipage}[b]{0.4\textwidth}
        \centering
        \includegraphics[width=\textwidth]{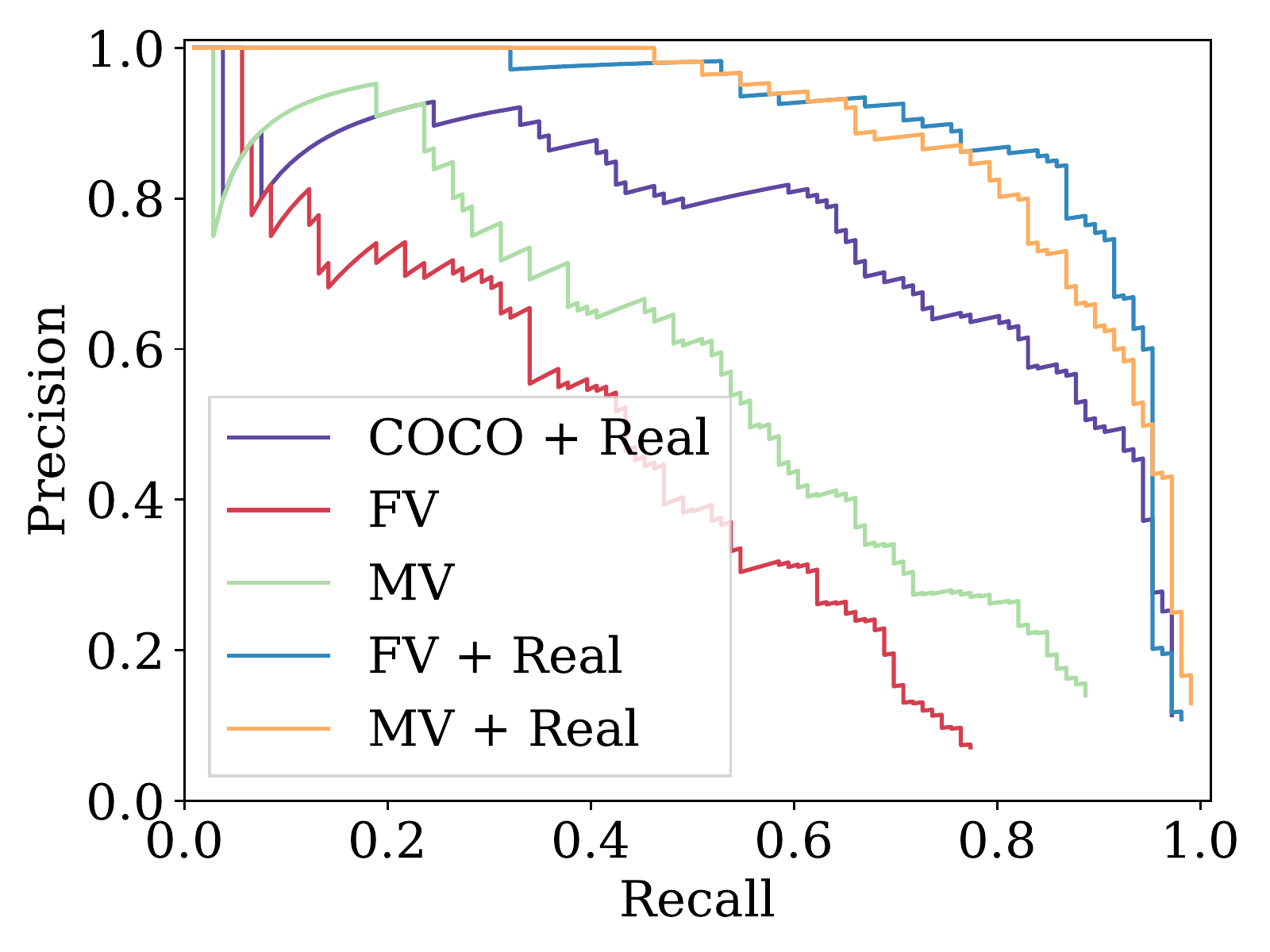}
        {\textbf{(c)} Box;}
        \label{fig:pr_30k_box}
    \end{minipage}
    \caption{
    Precision-recall curves of different variants of the detectors.
    For abbreviations refer to Fig.~\ref{fig:ap_30k}.
    }
    \label{fig:pr_30k}
\end{figure}

Fig.~\ref{fig:pr_30k} shows the precision-recall curves of different networks for each class.
Consistently, the networks trained on the fixed viewpoint dataset and fine-tuned on the real dataset out-perform other variations.
This trend is only less prominent in the case of \emph{sphere} class, where, seemingly, due to the smaller examples of this class in the real dataset~(Table~\ref{tab:imbalanced}) the training benefits less from fine-tuning on the real dataset for some values of iso-f1 surfaces.
This observation is also visible in Table~\ref{tab:ap_30k}.
We hypothesize that more real sphere examples could help the detector in improving the \ac{AP} score for spheres.

\subsection{Individual contribution of texture patterns}

A valid question in domain randomization research is the contribution of including various textures as well as the importance of sample sizes. To study this question, we have created smaller synthetic datasets with only $6k$ images, where in each of them one specific texture is missing.
Similar to previous subsection, MobileNet pre-trained on ImageNet was selected as the classifier \ac{CNN}, but the detectors were instead trained on these smaller synthetic datasets and then, fine-tuned on the real dataset.

The training of all networks on synthetic datasets lasted for 130 epochs, which was found to be the point where the \ac{mAP} did not improve over the validation set, with an exponentially decaying learning rate starting at $\alpha_0=0.004$, $k=0.95$ and $t=50k$ steps.
Finally, these networks were fine-tuned with the real-image dataset for 1100 epochs, with a constant learning rate $\alpha=0.001$.

\begin{figure}[!ht]
\centering
    \begin{minipage}[b]{0.46\textwidth}
        \centering
        \includegraphics[width=\linewidth]{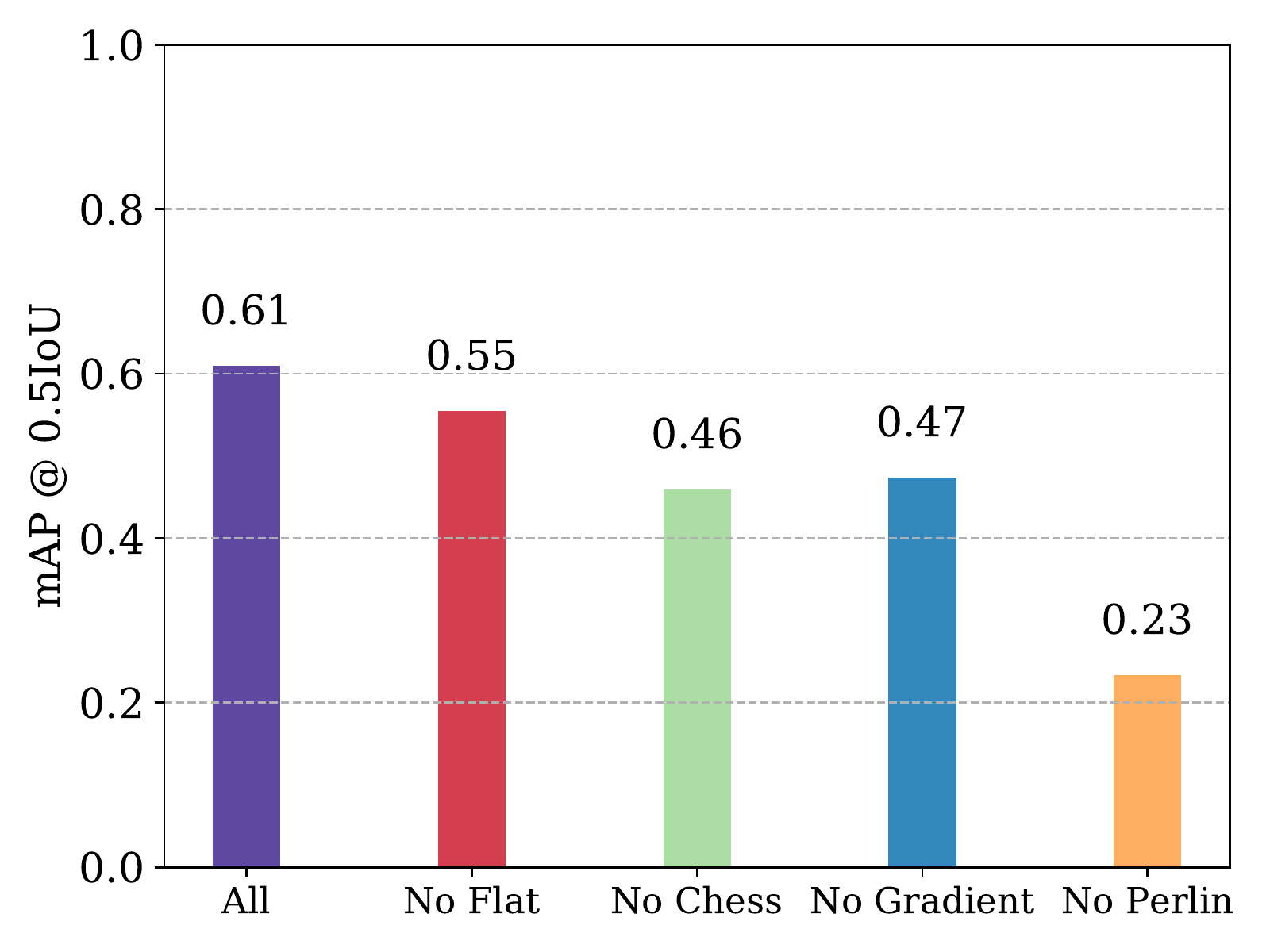}
        {\textbf{(a)} Before fine-tuning;}
        \label{fig:map_6k_no_real}
    \end{minipage}
    ~
    \begin{minipage}[b]{0.46\textwidth}
        \centering
    \includegraphics[width=\textwidth]{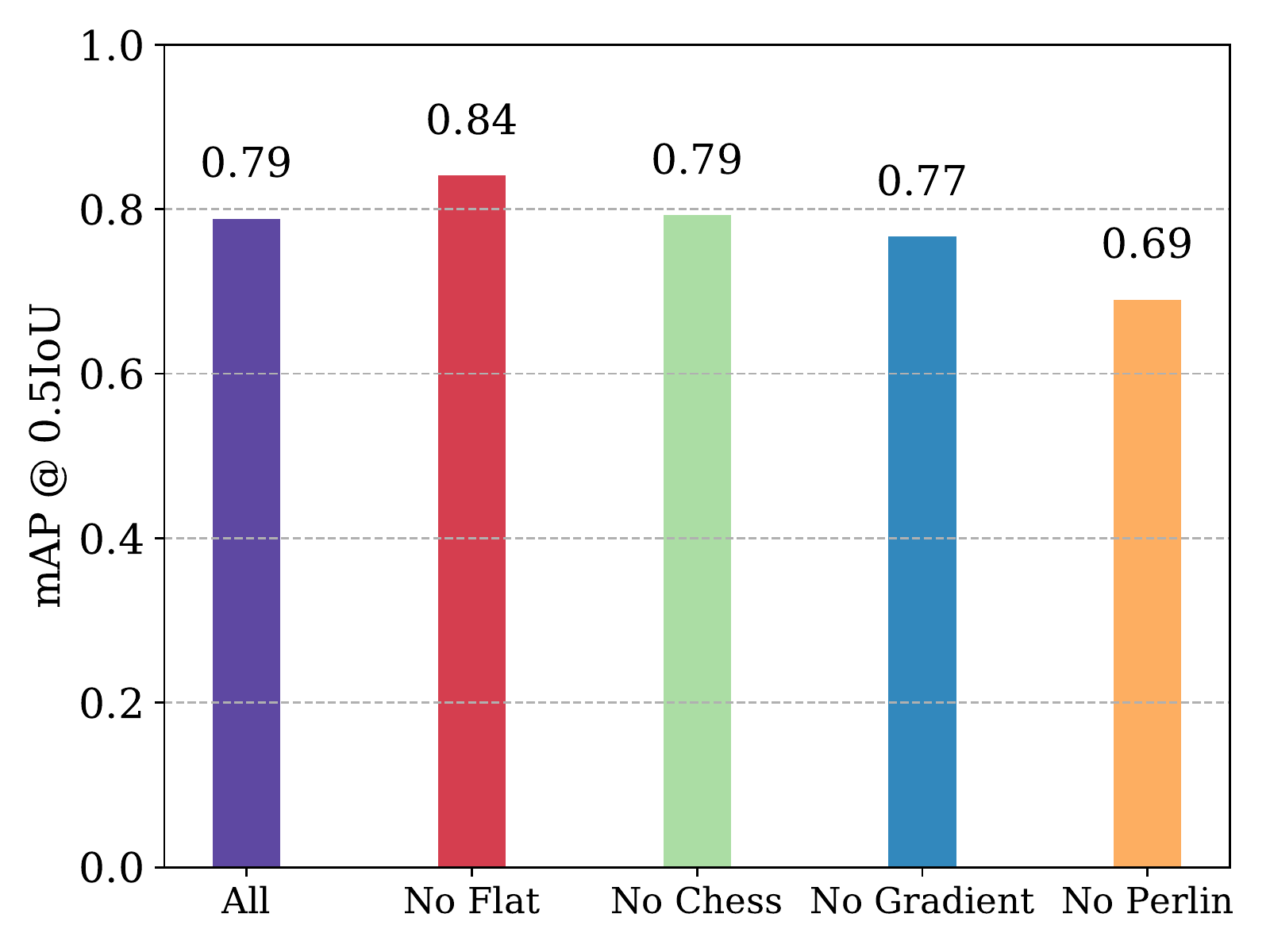}
        {\textbf{(b)} After fine-tuning;}
        \label{fig:map_6k_real}
    \end{minipage}
    \caption{
    Performance of \ac{SSD} on test set during training on smaller datasets of 6k images, each missing a type of texture, with the exception of the baseline, prior and after fine-tuning on real image dataset (\textbf{(a)},\textbf{(b)} respectively).
    }
    \label{fig:map_6k}
\end{figure}

\begin{figure}[!ht]
\centering
    \begin{minipage}[b]{0.4\textwidth}
        \centering
        \includegraphics[width=\textwidth]{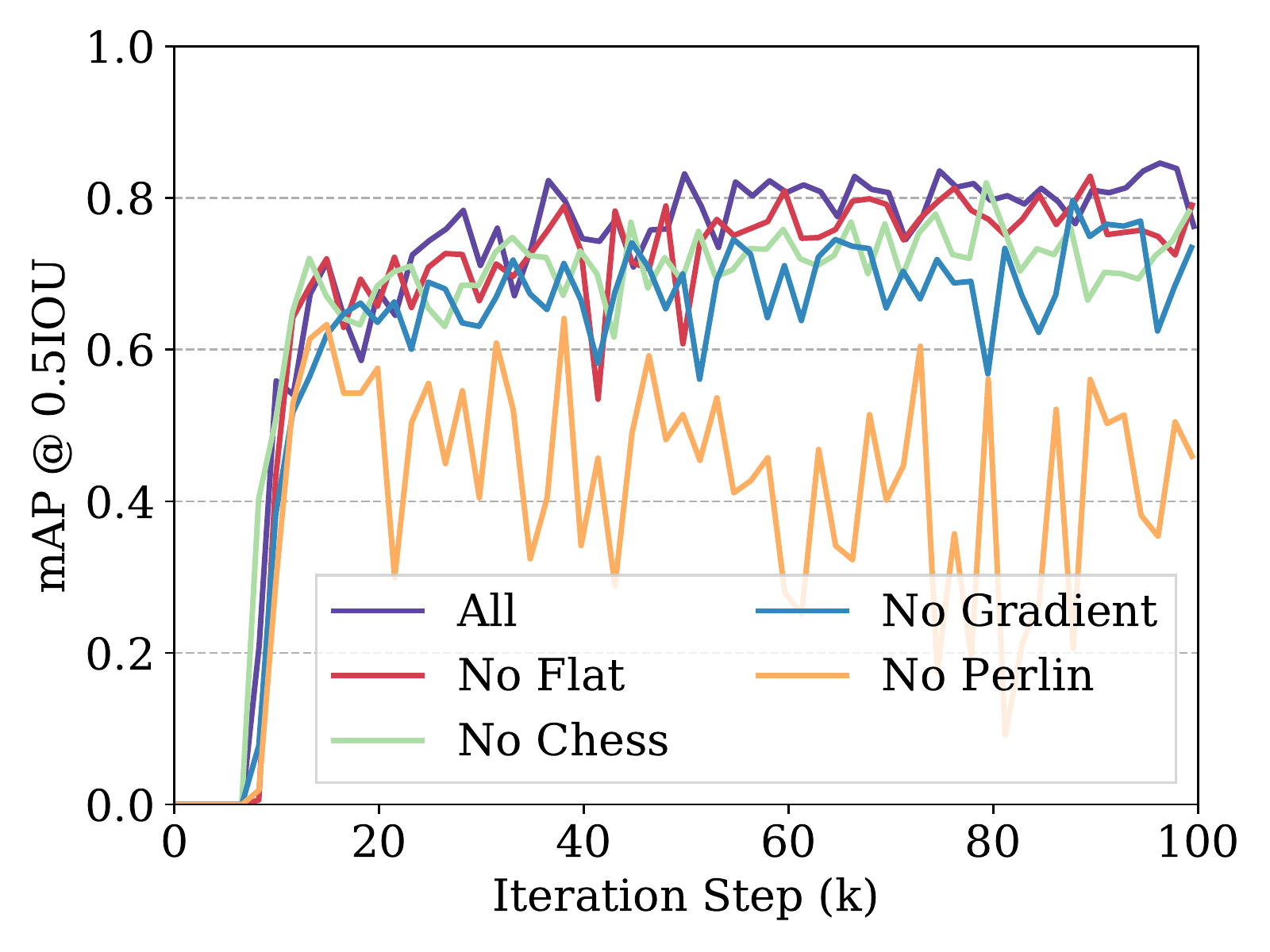}
        {\textbf{(a)} Before fine-tuning;}
        \label{fig:val_6k_no_fine_tune}
    \end{minipage}
    ~
    \begin{minipage}[b]{0.4\textwidth}
        \centering
    \includegraphics[width=\textwidth]{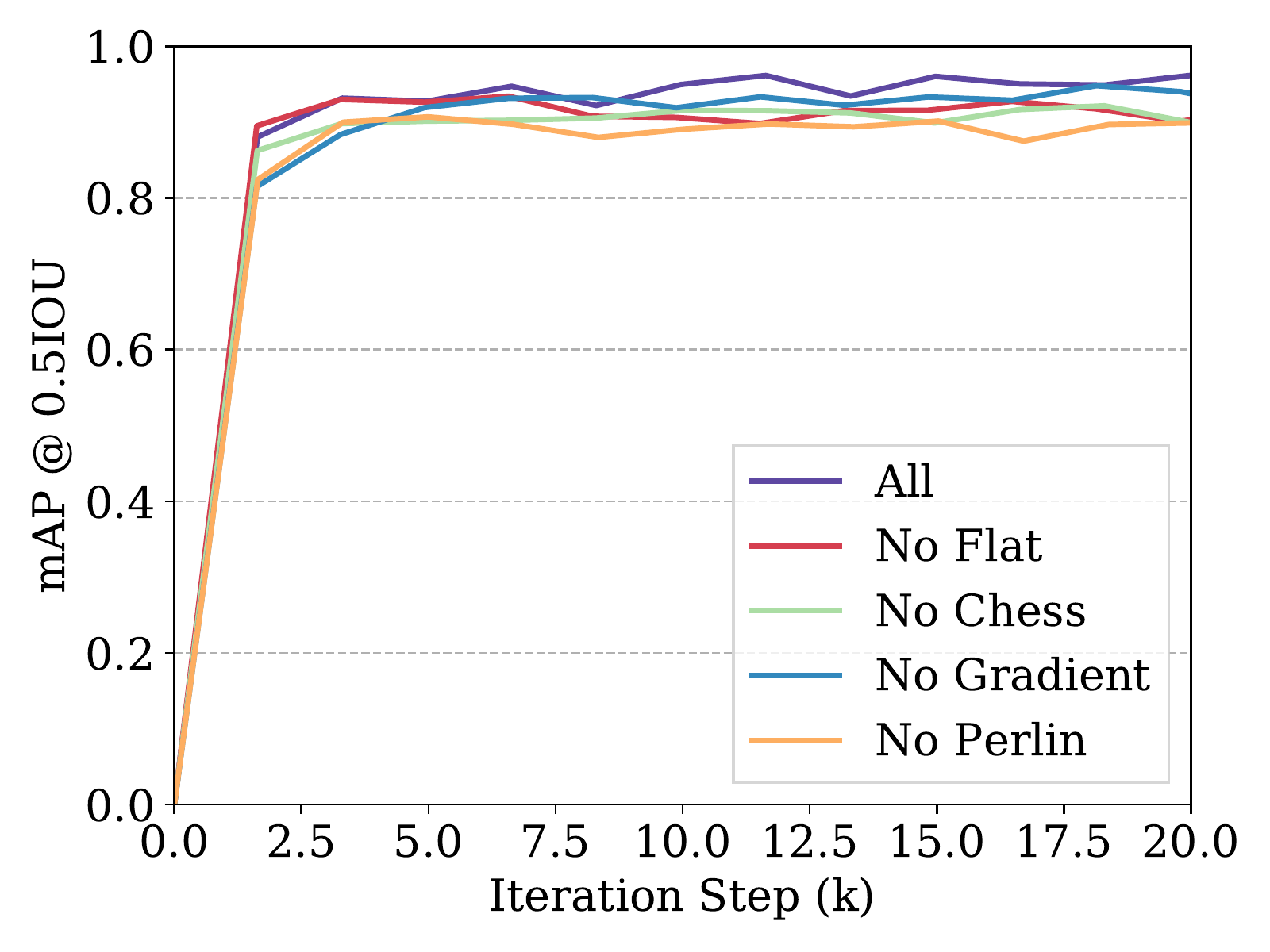}
        {\textbf{(b)} After fine-tuning;}
        \label{fig:val_6k_fine_tune}
    \end{minipage}
    \caption{
    Performance of \ac{SSD} on validation set during training on smaller datasets of 6k images, each missing a type of texture, with the exception of the baseline, prior and after fine-tuning on real image dataset (\textbf{(a)},~\textbf{(b)} respectively).
    }
    \label{fig:val_6k}
\end{figure}

\begin{figure}[!ht]
\centering
    \begin{minipage}[b]{0.4\textwidth}
        \centering
        \includegraphics[width=\textwidth]{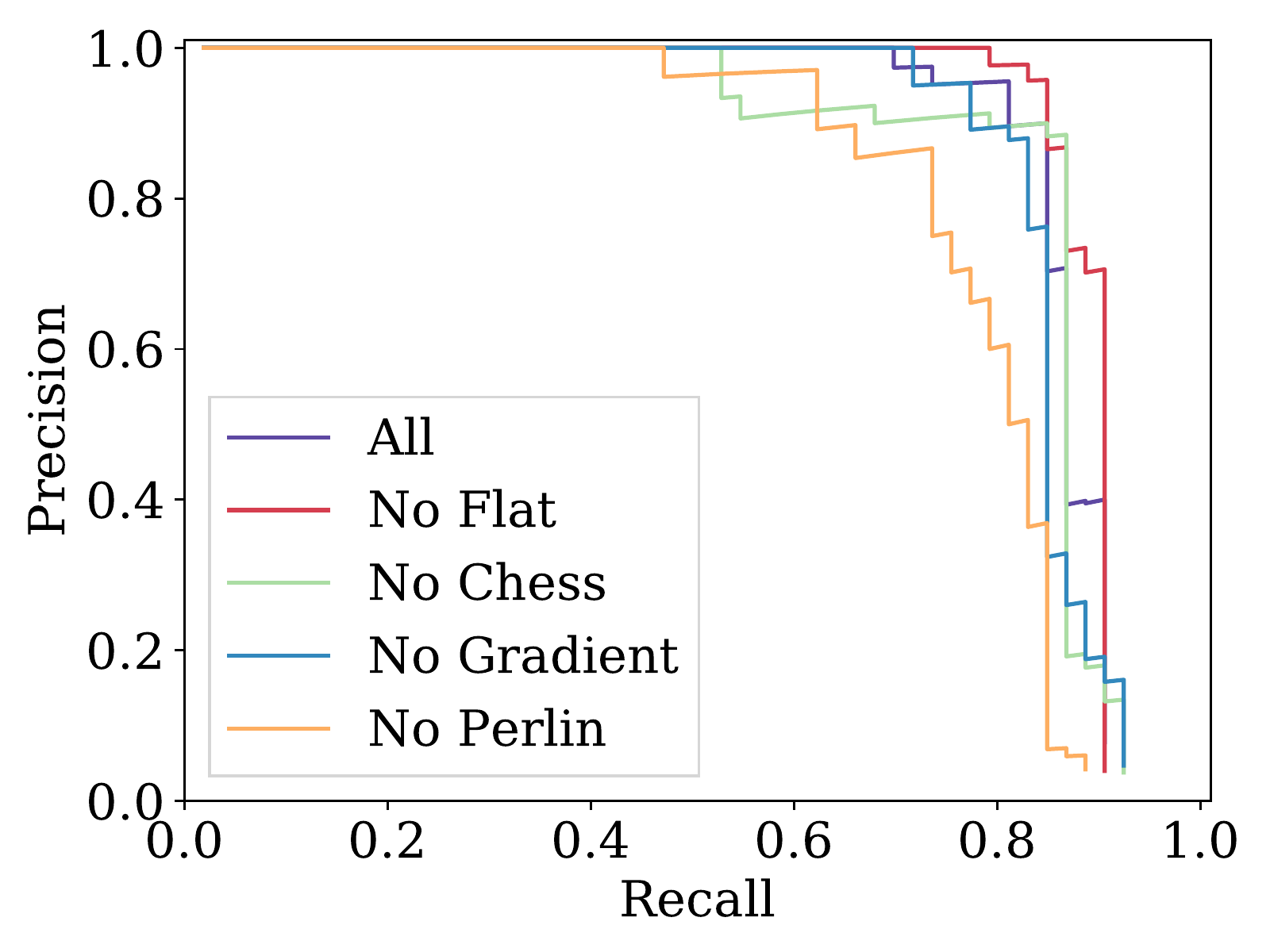}
        {\textbf{(a)} Sphere;}
        \label{fig:pr_6k_sphere}
    \end{minipage}
    ~
    \begin{minipage}[b]{0.4\textwidth}
        \centering
        \includegraphics[width=\textwidth]{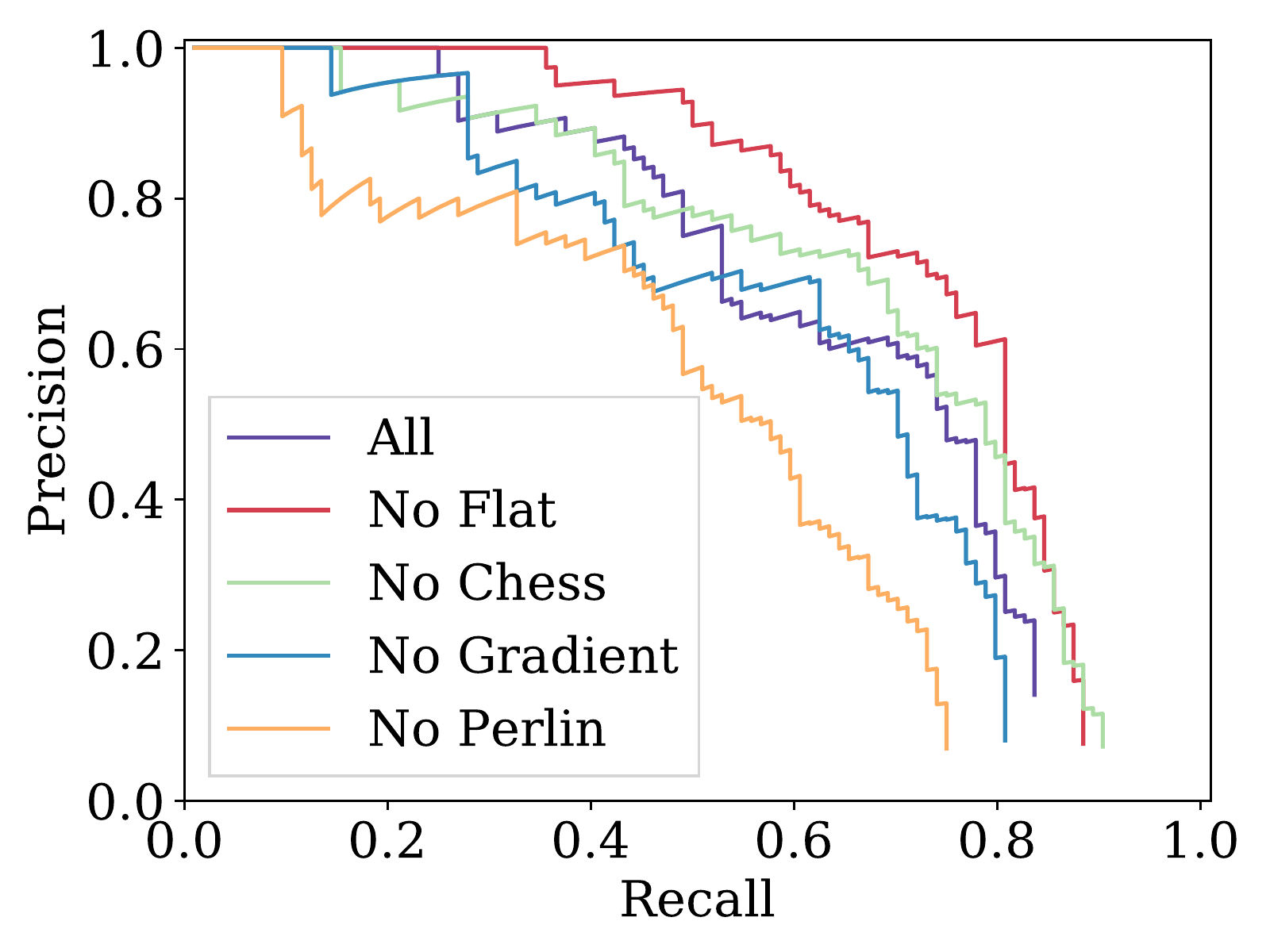}
        {\textbf{(b)} Cylinder;}
        \label{fig:pr_6k_cylinder}
    \end{minipage}
    ~
    \begin{minipage}[b]{0.4\textwidth}
        \centering
        \includegraphics[width=\textwidth]{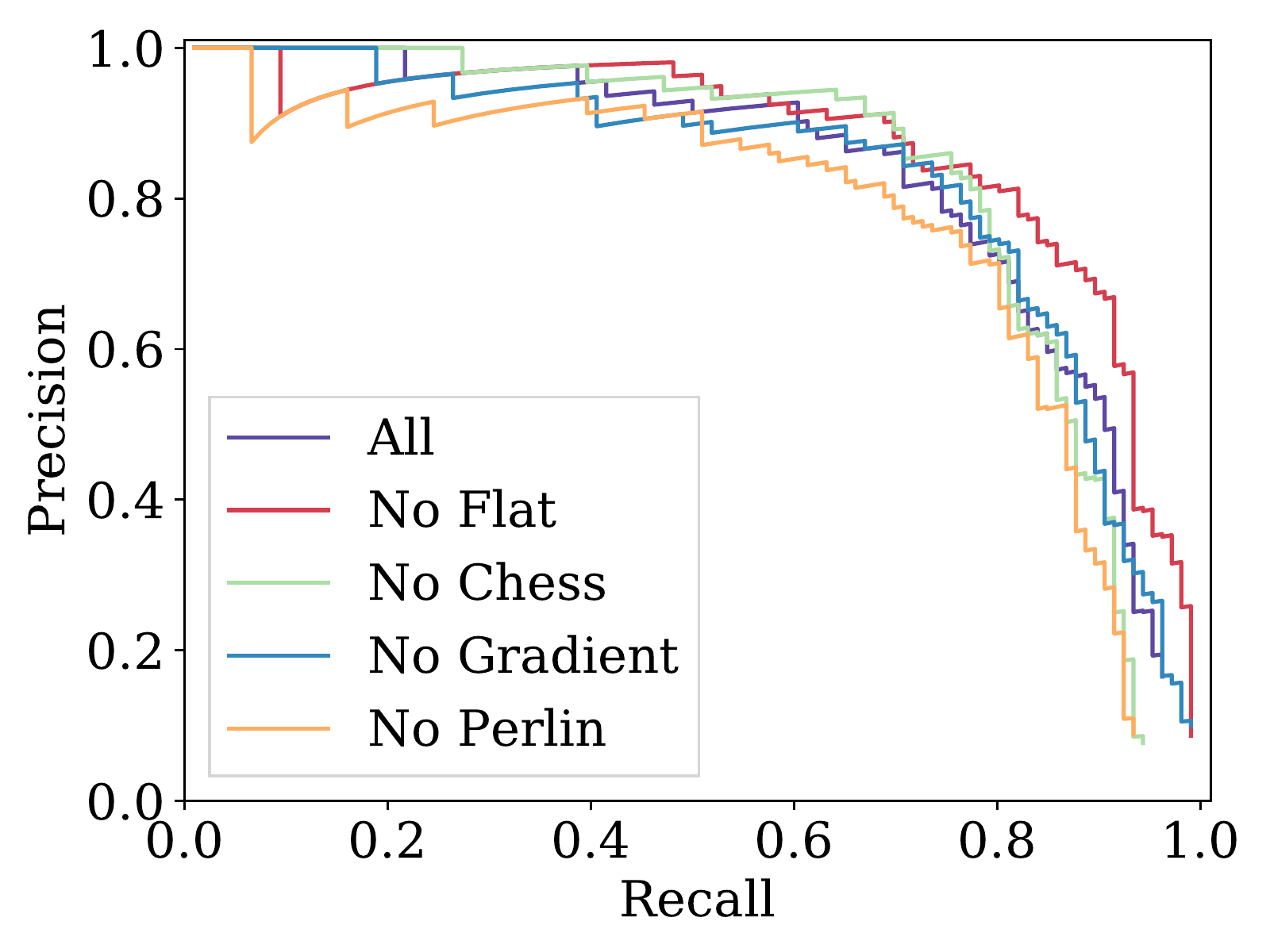}
        {\textbf{(c)} Box;}
        \label{fig:pr_6k_box}
    \end{minipage}
    \caption{
    Precision-recall curves of different variants of the detectors after fine-tuning on the real dataset.
    For abbreviations refer to Fig.~\ref{fig:ap_30k}.
    }
    \label{fig:pr_6k}
\end{figure}

The results of these experiments are reported in Fig.~\ref{fig:map_6k} and Table~\ref{tab:chunks}.
By comparing figures \textbf{(a)} and \textbf{(b)} in Fig.~\ref{fig:map_6k} it is clear that all variations have benefited from fine-tuning with the real dataset.

About the individual performances, generally speaking, by removing more and more complex textures~(flat to be the least complex and Perlin to be the most complex), the performance hurts, and we found Perlin noise to be a vital texture for object detection, while the flat texture has the least significance.
Consistent with this observation, according to Fig.~\ref{fig:map_6k}~\textbf{(b)}, the small dataset with all the textures cannot always compete with some of the datasets where a texture is missing.

According to Figures~\ref{fig:val_6k} and~\ref{fig:map_6k}~\textbf{(a)}, the detector trained on the small dataset with all texture classes, outperformed other variations on the \textbf{validation set} during training, however, presumably due to smaller number of samples and simultaneously, so many texture classes, over-fitted to the objects in the train set and failed to generalize as well as others to the objects in the test set.

Regarding the number of samples, our in house study with $200k$ synthetic images~(unreported) in line with the findings in~\cite{tremblay2018training}, suggests that more is not always better.
The network trained on our smaller dataset of only $6k$ images without the ``flat'' texture has even slightly out-performed the network that was trained on $30k$ synthetic images.
This result seems to be consistent for detectors with classifiers trained on real images, trained on synthetic data and then again fine-tuned with real samples.
After a fixed number of images, the \ac{mAP} performance oscillates for one or two percents.

\begin{table}[!ht]
\centering

\caption{\ac{SSD} performance on test set after train on each of the 6k sub--datasets and fine-tuned on real images.}
\label{tab:chunks}
\begin{tabular}{lllll}
\hline\noalign{\smallskip}
Training dataset $\qquad$ &
mAP $\qquad$ &
AP Box $\qquad$ &
AP Cylinder $\qquad$ &
AP Sphere $\qquad$ \\
\hline\noalign{\smallskip}
All         & 0.7885 & 0.8344 & 0.6616 & 0.8694 \\
No Flat     & \textbf{0.8410} & \textbf{0.8775} & \textbf{0.7546} & \textbf{0.8910} \\
No Chess    & 0.7925 & 0.8332 & 0.6958 & 0.8485 \\
No Gradient & 0.7668 & 0.8296 & 0.6172 & 0.8536 \\
No Perlin   & 0.6901 & 0.7764 & 0.5058 & 0.7880 \\
\hline
\end{tabular}

\end{table}

\section{Conclusions and future directions}
\label{sec:conclusions}

In this work we have shown that multi-category object detection pipelines can significantly benefit from pre-training on synthetic non-photo-realistic datasets.
Our modifications to an open-source plugin have enabled us to rapidly test different variations in the synthetic data and assess the importance of various components such as texture complexity and sample size.

According to our experiments, increasing texture complexity in the synthetic data should be compensated by larger number of samples, however, big gains in detector accuracy can be obtained with synthetic datasets that are orders of magnitude smaller than \ac{COCO} or ImageNet as long as a classifier trained on real datasets is being used.

Our modifications to the plugin for the synthetic data generation will facilitate the creation of scenes for other types of studies in \emph{domain randomization}, such as the impact of clutter and the increasing number of new object classes.
The choice of Gazebo will facilitate the creation of scenes for deep learning experiments in robotics, such as object tracking and mobile manipulation.  

In real scenarios where final performance metric is usually the most pertinent consideration, various data augmentation techniques such as color intensity distortions, random crops, etc. should be added to the training pipeline of domain randomization to improve the generalization capabilities of the detector at test time.

We believe enriching the plugin with more texture categories and combinations of categories can significantly improve the synthetic data quality for domain randomization studies.
More specifically, currently no synthetic object can have more than one texture, where as in reality, e.g., a box can have different textures at each side.
Another limitation of this plugin is that it not possible to stack objects on top of one another, however, in our test scenarios many objects were placed on top of each other.
Removing these limitations can widen the applicability of the plugin in different domain randomization scenarios.

Finally, with the advance of deep instance segmentation methods~\cite{he2017mask}, a similar study should be conducted to assess the applicability of domain randomization on object category segmentation.

\section*{Acknowledgements}

\hspace{\parindent}
This work is partially supported by the Portuguese Foundation for Science and Technology (FCT) project [UID/EEA/50009/2013].
Atabak Dehban and Rui Figueiredo are funded by FCT PhD grants PD/BD/105776/2014 and
PD/BD/105779/2014, respectively.
The Titan Xp GPUs used for this research were donated by the NVIDIA Corporation.


\bibliographystyle{splncs}
\bibliography{egbib}

\end{document}